# Navigating the Risks of Using Large Language Models for Text Annotation in Social Science Research

Hao Lin and Yongjun Zhang*

Department of Sociology and Institute for Advanced Computational Science, Stony Brook University, New York, United States

* Both authors contributed equally to this paper and are listed in alphabetical order. Direct correspondence to Yongjun Zhang; email: yongjun.zhang@stonybrook.edu

## Abstract

Large language models (LLMs) have the potential to revolutionize computational social science, particularly in automated textual analysis. In this paper, we conduct a systematic evaluation of the promises and risks associated with using LLMs for text classification tasks, using social movement studies as an example. We propose a framework for social scientists to incorporate LLMs into text annotation, either as the primary coding decision-maker or as a coding assistant. This framework offers researchers tools to develop the potential best-performing prompt, and to systematically examine and report the validity and reliability of LLMs as a methodological tool. Additionally, we evaluate and discuss its epistemic risks associated with validity, reliability, replicability, and transparency. We conclude with several practical guidelines for using LLMs in text annotation tasks and offer recommendations for more effectively communicating epistemic risks in research.

## Introduction

Coding complex concepts from text corpora is one of the primary tasks for both quantitative and qualitative social scientists. Although human coding remains the gold standard, computer-aided coding has demonstrated the capacity to efficiently perform sophisticated text annotation tasks (Nelson, Burk, Knudsen, & McCall, 2021). For instance, in social movement studies, scholars have devoted considerable resources to building protest event databases by leveraging supervised machine learning techniques to extract protest information from large-scale news articles and social media posts (H. Zhang & Pan, 2019; Hanna, 2017). Recently, many scholars have highlighted the potential of swift advancements in large language models (LLMs) to transform social science research by automating key stages of the research pipeline, including summarizing literature review, generating and testing hypotheses, and conducting data collection and analysis (Ziems et al., 2024; Messeri & Crockett, 2024; Manning, Zhu, & Horton, 2024; Bail, 2024). Yet, little research has been done to systematically examine the validity, reliability, transparency, and replicability of using LLMs to code complex concepts for social science research. LLMs may carry great epistemic risks when social scientists treat them as knowledge-production partners (Messeri & Crockett, 2024). To address this gap, this study asks the crucial question of whether, and under what circumstances we, as social scientists, can trust results produced by LLMs. Specifically, we investigate the risks of using LLMs for text annotation in social science research and how these risks can be better mitigated and communicated. In response, we explore and propose a framework for incorporating LLMs into text annotation practices in a manner that prioritizes reliability, validity, replicability, and transparency.

Computer scientists have demonstrated the potential of LLMs to achieve high accuracy by benchmarking their performance across a range of tasks, including multiple-choice questions (MMLU), reasoning (HellaSwag), coding (HumanEval benchmark), and mathematics (MATH benchmark). For instance, the multitask accuracy benchmark spans 57 tasks across STEM, humanities, social sciences, and other domains(Hendrycks et al., 2020). These evaluations provide the broader community with a snapshot of LLMs' knowledge acquisition and reasoning capabilities. For example, the average accuracy rates across benchmark datasets - including MMLU, HellaSwag, HumanEval, BBHard, GSM-8K, and MATH - are 79.45% for GPT-4, 80.08% for Gemini 1.5 Pro, and 84.83% for Claude 3 Opus [1]. In scientific contexts, LLMs have been lauded for their contributions to enhancing scientific discovery and objectivity, helping to overcome limitations related to scientists' time, attention, cognitive capacities, subjectivity, and bias (Messeri & Crockett, 2024; Ziems et al., 2024; Manning et al., 2024). However, LLMs also pose challenges to scientific understanding, introducing potential epistemic risks such as data leakage, hallucinations, strong alignment with prevailing social norms, and persistent issues with replicability, transparency, and interpretability.

In this paper, we use the field of social movement studies as an illustrative case to outline several risks associated with using LLMs, specifically OpenAI's GPT-4 and Meta's Llama3, for text annotation, and we propose approaches to mitigate some of the associated epistemic risks. We focus on typical tasks undertaken by social movement scholars when creating protest databases from newspaper articles, such as determining the relevance of articles to protests and extracting crucial information, including participant numbers, objectives, tactics, and involved social movement organizations. In the social movement literature, a frequently used benchmark dataset for analyzing protests in the United States is the Dynamics of Collective Action (DoCA) project (Earl, Martin, McCarthy, & Soule, 2004; Hanna, 2017). This project annotated news articles from the New York Times (NYT) covering the period 1960 - 1995 to construct a detailed event database containing information at both the article and event levels, such as the reporting year, article title, size of protests, claims, social movement organizations, locations, timing, and more [2]. Because the DoCA dataset does not include the original NYT articles, we align the annotated NYT corpus (1987-2007) from the Linguistic Data Consortium (LDC) with the DoCA dataset and employ GPT-4 and Llama3 to code various protest-related concepts. We conduct a series of experiments to evaluate the validity, reliability, transparency, and replicability of the outputs generated by LLMs and propose a workflow for using LLMs in text annotation that prioritizes the rigorous evaluation and reporting of these key dimensions.

[1] For more updated model comparisons, visit: `https://www.vellum.ai/llm-leaderboard`

[2] Further details can be found here: `https://web.stanford.edu/group/collectiveaction/`

We compare the outputs of LLMs (gpt-4-turbo, Llama-3-8B, and Llama-3.3-70B) against human-coded results from the DoCA project. Overall, we find that LLM performance declines as coding task complexity increases. GPT-4 performs well on simpler tasks, such as binary classification (e.g., identifying police presence), but shows relatively poor performance on multilabel and multiclass classification tasks. Our results also indicate that the reliability of LLM outputs varies with task complexity, exhibiting higher reliability for simpler tasks compared to more complex ones. Importantly, we demonstrate that social scientists can prompt LLMs to generate intermediate reasoning steps, thereby enhancing the transparency of coding decision-making processes.

We also stress that scholars must exercise caution when using LLMs as primary coders. We propose that to achieve high levels of validity and reliability, scholars need to test the accuracy and stability of the prompts, identify the best prompting strategy using a small development dataset, and report performance metrics on a randomly selected small evaluation dataset with human evaluations after scaling the prompts to the entire dataset. Meanwhile, LLMs can serve as secondary coding assistants by producing step-by-step reasoning results and suggesting preliminary classifications to aid human coders. We use OpenAI's GPT-4 and Meta's Llama3 models as key examples; however, the guidelines we propose and the epistemic risks we identify apply broadly to the use of other LLMs as well.

There are several important caveats when interpreting our results. First, we only experimented with zero-shot text classification using different prompting strategies. Scholars can further explore alternative strategies to use LLMs such as few-shot prompting or even fine-tuning domain-specific LLMs. Second, given the rapid advancement of LLMs, some of the limitations and risks discussed in the paper may be mitigated in the near future. Third, we focus on a specific domain of social movement studies, and the performance of LLMs may vary across different research areas. Finally, due to data safety and copyright concerns, we encourage prioritizing open-source models (e.g., Meta's Llama, DeepSeek R1) over proprietary ones (e.g., OpenAI's GPT). This is especially viable when sufficient computing resources exist. In this paper, we conducted experiments with gpt-4-turbo (proprietary), Llama-3-8B(open-source), and Llama-3.3-70B (open-source) as leading representatives of each model type.

# The Adoption of LLMs for Text Annotation

Transformer-based LLMs, such as BERT, RoBERTa, and Longformer, have been widely adopted by social scientists for textual analysis tasks, particularly in contexts where training datasets are relatively limited, by leveraging the advantages of transfer learning (Vaswani et al., 2017; Devlin, Chang, Lee, & Toutanova, 2018; Liu & Salganik, 2019; Beltagy, Peters, & Cohan, 2020; Do, Ollion, & Shen, 2022; Wankmüller, 2022). Since social scientists often face challenges in labeling large-scale training examples necessary to build a model from scratch, transfer learning offers an effective solution, allowing scholars to apply knowledge acquired during the pretraining stage on other data sources to their specific tasks in the target domain. That is, social scientists can leverage transformer-based models pretrained on large-scale textual or multimodal datasets and then fine-tune them for downstream tasks, even when only a limited number of labeled examples are available.

The typical approach involves (1) selecting an appropriate model, such as BERT, which is pretrained on massive corpora including Wikipedia and the Google Books Corpus, and (2) fine-tuning the model parameters using human-annotated training examples for domain-specific tasks (e.g., binary or multiclass classification). Scholars have applied this strategy to study a variety of topics, such as populism (Bonikowski, Luo, & Stuhler, 2022), Sinophobia (Y. Zhang, Lin, Wang, & Fan, 2023), and public sentiment on social media (Lu, Pan, & Xu, 2021). In the study of populism, Bonikowski et al. (2022) used RoBERTa to classify six subtle and rare political frames. Their findings demonstrated that fine-tuning pretrained transformer-based models is more effective than other classification algorithms such as random forest. However, as shown in their paper, they manually annotated 2,224 paragraphs to create the training dataset, and to address the challenges of an imbalanced dataset, they employed hybrid active learning to iteratively train the model. It required 11 steps to fine-tune the model to achieve satisfactory performance. This case illustrates that fine-tuning transformer-based language models for domain-specific tasks and achieving reliable performance demands considerable expertise in machine learning and programming, which continues to impede their broader adoption in the social sciences.

It is important to note that LLMs for automated textual analysis are not the only NLP techniques adopted by social scientists. The choice of method should align closely with the specific research question, as well as broader methodological considerations such as model transparency and interpretability. Other widely used NLP techniques, such as word embeddings and topic modeling, each offer distinct methodological advantages and should be prioritized when appropriate to avoid over-engineering or unnecessary complexity (DiMaggio, 2015; Grimmer, Roberts, & Stewart, 2022). For instance, word embeddings convert words or documents into numerical representations, enabling the analysis of semantic similarities and semantic networks of entities (Kozlowski, Taddy, & Evans, 2019; Arseniev-Koehler, 2024; Garg, Schiebinger, Jurafsky, & Zou, 2018). Topic modeling discovers recurring themes within a collection of documents by identifying clusters of co-occurring words, allowing social scientists to analyze discourse patterns, track issue salience, and study the evolution of ideas over time(DiMaggio, Nag, & Blei, 2013; Nelson, 2020). Although tasks such as topic modeling can be reframed into LLM-based approaches, we deliberately avoid doing so because traditional topic models offer greater transparency through their explicit probabilistic structures -making visible how topics are composed of word distributions and how documents are associated with multiple topics (Blei & Lafferty, 2009; Roberts et al., 2014). Moreover, given the distinct applications of each technique, there is a well-established body of literature guiding social scientists on how and when to employ various automated textual analysis methods, whether by integrating them with traditional qualitative approaches (Nelson, 2020) or combining multiple methods within an iterative research framework (Grimmer et al., 2022).

The field of NLP has recently evolved from models primarily designed for understanding input texts, such as BERT, RoBERTa, and Longformer, to models focused on next-token prediction, such as GPT and Llama. The rapid advancement has transformed computational social science, enabling scholars to use GPT-based models for zero-shot or few-shot text classification without requiring prerequisite fine-tuning expertise (Brown et al., 2020; Chae & Davidson, 2023; Ziems et al., 2024; Do et al., 2022). LLMs with text generation capabilities are trained on massive corpora of textual data and are capable of producing human-like text. These models, with billions of parameters, learn the probability distribution of tokens in the training data, allowing them to perform out-of-sample predictions and generate coherent text based on an input sequence of tokens or a given prompt.

Scholars are still debating whether social scientists should fine-tune transformer-based language models for information retrieval with transfer learning or simply use LLMs with generation capabilities such as OpenAI's GPT models for zero-shot or few-shot learning. Although limited research suggests that the latter approach does not always outperform fine-tuned models in text classification, LLMs such as GPT models offer a distinct advantage in zero-shot and few-shot learning, with the added capability of generating higher-quality explanations than those typically provided by crowd workers (Chae and Davidson 2023; Ziems et al. 2023). A recent study shows that OpenAI's GPT models outperforms crowd workers in annotation tasks based on four samples of tweets and news articles, achieving 25% higher zero-shot accuracy and being thirty times cheaper than MTurk (Gilardi, Alizadeh, & Kubli, 2023). Notably, LLMs are also widely used by crowd workers on MTurk to automate text annotation tasks, with an estimated prevalence of 30% (Veselovsky, Ribeiro, & West, 2023).

# A Framework for Text Annotation Using LLMs

Given the growing use of LLMs for text annotation, it is essential for social scientists to evaluate workflows and establish standards to guide these processes. We propose that LLMs can serve two distinct roles in text annotation: (1) as primary coders, with humans evaluating their performance for large-scale data, or (2) as secondary assistants, with humans acting as the final decision-makers for small- to medium-scale data. Figure 1 presents a workflow for using LLMs in text annotation, highlighting these two primary roles.

For social scientists, the appropriate role of LLMs - whether as primary coders or secondary assistants - depends on several factors, including dataset size, task complexity (e.g., binary versus multiclass classification), and acceptable error tolerance (e.g., required accuracy rates). If researchers are dealing with a large-scale dataset with low task complexity and relatively high error tolerance, LLMs can serve as the primary coder role while humans take on the evaluation role. To ensure high-quality text annotation, first we need to develop a coherent and consistent codebook, with all measured concepts clearly defined. Second, researchers need to build a development dataset by randomly selecting 100 $\sim$ 200 cases to iteratively test various prompting strategies to determine the best-performing prompt. Third, with the best-performing prompt we apply it to the entire dataset to generate annotation outputs. Finally, researchers need to further construct a separate evaluation dataset by randomly selecting additional 100 $\sim$ 200 cases to report performance metrics such as accuracy, precision, recall, and F1 score. Thus, the whole process takes both prompt engineering and data preparation into consideration.

Identifying the best-performing prompt is critical for achieving the desired outputs in this process, yet it remains a challenging task. Prompt selection is inherently iterative and experimental, relying on trial and error, as no single optimal prompt exists for every task or model. Moreover, the theoretically infinite space of possible prompts and the reality that prompts optimized for one model may not perform equally well on newer versions further complicates the search for an optimal prompt for a given task. To address this challenge, the most practical approach by far is to adopt established prompting strategies and rigorously evaluate them on a high-quality, human-labeled development dataset. The goal is to identify a prompt that consistently delivers better performance, such as higher accuracy and greater robustness, compared to alternative options. This approach aligns with best practices in scientific applications of LLMs across domains such as healthcare (L. Wang et al., 2024; Hu et al., 2024), language toxicity detection (Y.-S. Wang & Chang, 2022), and education (Lee et al., 2024).

The development and evaluation sets are archetypes we adapted from machine learning. The development set, also known as the validation set, is used in supervised machine learning to fine-tune model hyperparameters and select the optimal model. In our context, it is similarly used to test and determine the best-performing prompt. The evaluation set, sometimes also known as the test set, is used to evaluate the performance of the model. Here, it is also used to evaluate the performance of the best-performing prompt and the chosen model. Both datasets are sampled separately and randomly from the larger dataset, ideally ensuring they are independent of each other. This separation is crucial, as we cannot guarantee that an LLM will not incorporate examples seen during the development stage into its outputs when used interactively. By maintaining independence between these sets, we reduce the risk of artificially inflated performance estimates. The size of the development and evaluation sets is flexible and depends on time and budget constraints, and we recommend sampling at least $100 \sim 200$ cases for each set. Both datasets require human labeling and serve as the ground truth for the development and evaluation stage accordingly.

However, if the annotation task is complex, requires strong reasoning skills, the data size is manageable within the budget, the error tolerance is low, or the analytic method is inductive, then LLMs can act as coding assistants. In this role, they help with text summarization and provide preliminary reasoning for annotations. Similar to computer-assisted coding, LLM-driven coding enables scholars to efficiently digest key information from text or multimodal data (e.g., images and audio) and make final coding decisions.

# Prompt Engineering for Text Annotation Tasks

Given that the outputs of LLMs highly depend on the prompts developed by scholars for specific tasks, we further discuss common prompting strategies proposed by computer scientists and engineers. A prompt is a sequence of text input designed to direct LLMs to perform specific tasks, allowing scholars to bypass the need for fine-tuning LLMs for downstream tasks. Prompt engineering involves refining these prompts to steer LLMs toward producing desirable outputs. Various strategies for prompt engineering have been developed, such as in-context learning, role-play prompting, chain-of-thought prompting, and tree-of-thought prompting, to optimize the effectiveness of LLMs.

In-context learning is a prompting strategy that provides LLMs with a few input-output ex-amples to learn through analogy. Scholars create prompts by incorporating several examples as a demonstration context, enabling LLMs to discern hidden patterns in these examples and accurately perform downstream tasks. Notably, in-context learning does not update the LLMs' parameters, which significantly reduces computational costs and makes it easily applicable to social science problems (Dong et al., 2023). For instance, to classify whether a news article is related to a protest, social movement scholars can include several protest-related articles as contextual examples when formulating the prompt.

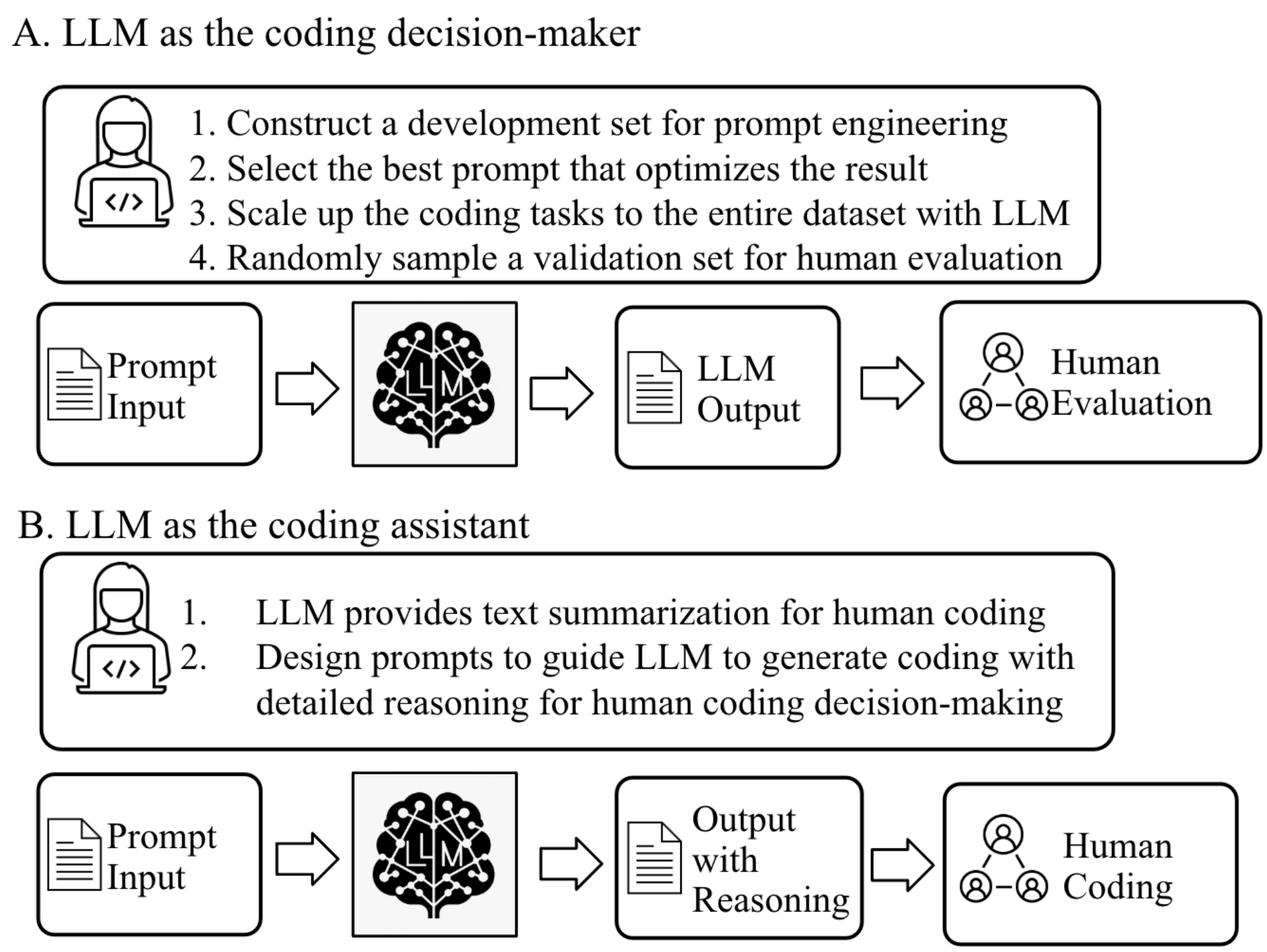

Figure 1: The Workflow of Using LLMs for Text Annotation. Note: the LLM icon was created by GPT-4o.

Role-play prompting is a novel strategy that assigns a specific role to the LLM, enabling it to produce outputs tailored to the identity and persona defined by the researcher (Kong et al., 2024). For instance, researchers can assign the role of an expert social movement scholar who excels at extracting information about protests from news articles to the LLM.

The chain-of-thought (CoT) prompting strategy directs LLMs to generate intermediate reasoning steps, improving their ability to perform complex reasoning tasks (Wei et al., 2022). CoT allows LLMs to break down multi-step problems into several intermediate steps, offering an interpretable window into the models' decision-making processes. CoT can be combined with in-context learning by including several examples (few-shot CoT) or activated by simply adding *'let us think step by step'* at the end of the prompt (zero-shot CoT) (Wei et al., 2022; X. Wang et al., 2023; B. Wang et al., 2023).

Instead of modeling the human thought process as a linear chain, the Tree-of-Thoughts (ToT) approach decomposes multi-step problems into a tree structure, where each thought represents an intermediate step in problem-solving (Yao et al., 2023; Long, 2023). This method allows language models to self-evaluate multiple intermediate thoughts through a deliberate reasoning process (Yao et al., 2023). Hulbert (2023) applies the ToT framework to guide LLMs in evaluating intermediate thoughts within a single prompt. For instance, the sample prompt used by Hulbert (2023) to trigger ToT is as follows: *"Imagine three different experts are answering this question. All experts will write down one step of their thinking, then share it with the group. Then all experts will proceed to the next step, and so on. If any expert realizes they're wrong at any point, they will leave. The question is..."*

Next, we present several examples demonstrating how these common prompting strategies can be applied to text annotation tasks in social science research, with a particular focus on social movement studies.

**Standard Zero-Shot Prompting with Role Play**

**System:** *You are an excellent AI language model trained to extract protest information from news articles.*
**User:** *You need to determine whether it discusses peace movement in the following article: <fulltext >. Please structure your response as follows: <numeric answer (Yes=1; No=0) >*
**Assistant:** *1*

**Zero-Shot CoT Prompting with Role Play**

**System:** *You are an excellent AI language model trained to extract protest information from news articles.*
**User:** *You need to determine whether it discusses peace movement in the following article: <fulltext >. Let us think step by step. Please structure your response as follows: <numeric answer (Yes=1; No=0) <Step-by-step reasoning >>*
**Assistant:** *1;*
*The article mentions the arrest of eleven people, including antiwar activist Philip Berrigan, who were part of a demonstration against war. They scaled a fence at the White House and threw red ink into a fountain as a form of protest.*
*The demonstrators are described as being affiliated with an antiwar group, specifically named the D.C. Coalition Against the U.S. in the Middle East.*
*This indicates that the protest was related to a peace movement, as it involved a group actively opposing war in the Middle East.*

**Zero-Shot ToT Prompting with Role Play**

**System:** *You are an excellent AI language model trained to extract protest information from news articles.*

**User:** *Imagine three different social movement experts are answering this question. All experts will write down one step of their thinking, then share it with the group. Then all experts will proceed to the next step, and so on. If any expert realizes they are wrong at any point, they will leave. The question is that You need to determine whether the following article discusses peace movement: <fulltext >. Please structure your response as follows: <Step-by-step reasoning ><numeric answer (Yes=1; No=0) >*

**Assistant:**

*Step 1: Identify the main activity described in the article.*

*- Expert 1: The article describes a group of people, including a known antiwar activist, who were arrested for scaling a fence at the White House and throwing red ink into a fountain.*

*- Expert 2: The demonstrators are affiliated with an antiwar group, indicating their primary motive is related to opposing war.*

*- Expert 3: The act of throwing red ink to simulate blood suggests a symbolic protest against violence or war.*

*Step 2: Determine the nature of the group involved in the activity.*

*- Expert 1: The group is identified as the D.C. Coalition Against the U.S. in the Middle East, which clearly positions itself against U.S. policies or actions in the Middle East.*

*- Expert 2: The involvement of Philip Berrigan, a notable antiwar activist, supports the interpretation that this is an antiwar protest.*

*- Expert 3: The name of the group and the nature of the protest both align with typical peace movement activities, which often involve symbolic actions to draw attention to the costs of war.*

*Step 3: Assess if the described actions align with those typical of a peace movement.*

*- Expert 1: Peace movements often use dramatic, non-violent actions to make a political point, which is consistent with the described protest.*

*- Expert 2: The use of red ink as a symbol for blood is a strong visual metaphor against war violence, aligning with peace movement strategies.*

*- Expert 3: The target of the protest, the White House, is a strategic location for making a bold statement against national policies on war.*

*Conclusion:*

*- Expert 1: The article describes a protest by an antiwar group using symbolic non-violent action, typical of peace movements.*

*- Expert 2: The involvement of known peace activists and the nature of the demonstration clearly point to a peace movement.*

*- Expert 3: The overall context and details provided in the article align with the characteristics of a peace movement protest.*

*Numeric Answer:1*

# Limitations of Using LLMs for Text Annotation

In this section, we examine the limitations of using LLMs for text annotation. It is important to note that we are not AI-Luddites; as advancements continue, we anticipate improvements that may mitigate these limitations and challenges, enabling more effective and ethical use of LLMs in social science research. Using social movement studies as an illustrative example, we highlight potential epistemic risks associated with employing LLMs for text annotation. These risks are particularly critical when scholars rely on LLMs as primary coders rather than as secondary coding assistants. Drawing on studies that focus on the risks and limitations of LLMs, we discuss strategies for incorporating these considerations into the prompt development process to mitigate such risks. Regardless of annotation task complexity, researchers should pay close attention to the epistemic risks associated with using LLMs and implement mitigation strategies to enhance validity, reliability, replicability and transparency.

## Are LLMs text annotation results valid?

Integrating LLMs into social science research, particularly for constructing protest event datasets in social movement studies, presents unique challenges. Movement scholars require high levels of coding accuracy due to the complexity and nuance of protests. A primary concern among social scientists is whether LLMs can achieve human-expert level precision in coding complex concepts, given their susceptibility to hallucinations and biases (Srinivasan & Chander, 2021; Bail, 2024). Recent studies suggest that LLMs, such as ChatGPT and similar technologies, can outperform crowd workers on tasks such as text summarization (Gilardi et al., 2023). However, a notable gap remains in achieving expert-level accuracy for domain-specific tasks. To address this, we benchmark LLM-produced results against the DoCA-LDC dataset to systematically assess whether LLMs can meet the standards of human-expert coding.

Previous research has shown that measurement validity in computational textual analysis tends to decline as the task complexity increases (Baden, Pipal, Schoonvelde, & van der Velden, 2022). Given the vast heterogeneity of textual documents and the diverse constructs social scientists examine, task complexity is often poorly operationalized and depends on multiple factors, including (1) the intricacy of the social scientific constructs, (2) the linguistic sophistication of the text, and (3) the relationship between the constructs and the textual data. Tasks become relatively more challenging when the constructs are modular and vaguely defined, the text is complex, and the relevant information is either ambiguously distributed across the text or requires external contextual knowledge to interpret (Do, Ollion, & Shen, 2024; Baden et al., 2022). Moreover, for machine learning classification tasks, task complexity generally increases along the following hierarchy: binary classification (two mutually exclusive classes), multiclass classification (more than two mutually exclusive classes), multilabel classification (multiple non-exclusive labels), and hierarchical classification (structured class relationships). In our study, we examine different levels of task complexity, including: (1) binary classification - the variable *police presence*, where relatively clear indicative information can be found in the text; (2) multiclass classification - the variable *protest size*, where indicative information is often less clearly defined; and (3) multilabel classification - *protest* *activities*, where key information may be dispersed across the text and multiple activities frequently occur within a single event.

For demonstration, we only use several common prompting strategies, including standard zero-shot with role-play, zero-shot CoT, and zero-shot ToT (see Appendix A for example prompts). These strategies are tested by querying the OpenAI GPT-4-turbo model using the matched DoCA-LDC dataset. To demonstrate the potential variations in the relationship between the performance of prompting strategies and task complexity, we conducted experiments using the same prompt question while incorporating different triggers to initiate various strategies. In the practical application, researchers should construct a development set to evaluate the performance of different strategies and select the best-performing one. Table 1 presents the accuracy and F1 scores across various text classification tasks.

*Police presence* is a binary variable indicating whether police involvement was mentioned at the event. *Protest size* is a categorical variable scaled from 1 to 6, where a score of 1 denotes a small group of 1-9 individuals, and a score of 6 corresponds to gatherings of 10,000 or more participants. *Protest activity* involves a multilabel classification problem where LLMs are tasked with identifying up to four activities from a predefined list, which includes bannering, candle-lighting, petitioning, holding signs, camping, sit-ins, physical attacks, looting, and more. To assess classification performance on protest activities, we adopted the *subset accuracy* metric. This metric is defined as the proportion of instances where all relevant protest codes listed in DoCA are completely encompassed within gpt-4-turbo's predictions. Consequently, a prediction is considered accurate only if every protest activity code generated by gpt-4-turbo for an article is inclusive of all corresponding codes in DoCA.

Our results show that the coding performance of gpt-4-turbo varies depending on the level of task complexity and the prompting strategies used. Overall, gpt-4-turbo performs well on less complex tasks but struggles with more complex tasks, such as multiclass and multilabel classification. It is important to note that we only conducted experiments using zero-shot prompts; results may differ when using other prompting strategies, such as few-shot prompts.

| Task Complexity | Zero-shot | Zero-shot CoT | Zero-Shot ToT |
|---|---|---|---|
| Binary (Police) | 0.84, 0.63 | 0.85, 0.64 | 0.82, 0.61 |
| Multiclass (Size) | 0.50, 0.46 | 0.48, 0.45 | 0.46, 0.43 |
| Multilabel (Activity) | 0.55, - | 0.41, - | 0.56, - |

Table 1: GPT-4-turbo Performance across Different Prompting Strategies and Tasks. All methods are with role play prompting.

## Are LLMs text annotation results reliable?

Another widely shared concern among scholars is the reliability of using LLMs in social science research. A critical question is whether LLMs produce consistent results when using the same prompting strategy and parameter settings. Our proposed approach involves running the same prompt multiple times (at least 25) on a development dataset and calculating the intraclass correlation (ICC) to measure inter-rater consistency. Each run is treated as a separate rater. According to Koo and Li's guideline (2016), ICC values below 0.5 indicate poor reliability, values between 0.5 and 0.75 suggest moderate reliability, values between 0.75 and 0.9 indicate good reliability, and values above 0.90 indicate excellent reliability.

To demonstrate, we used the OpenAI gpt-4-turbo model to classify a subset of the DoCA-LDC dataset, focusing on two specific tasks: police presence and protest size. To mimic a development set, we randomly selected 200 articles and queried the OpenAI API 25 times using the same prompt and parameter settings. Figure  illustrates the trend of average accuracy across these 25 runs. Overall, the accuracy scores fluctuate for both measures when repeated 25 times. We also use intraclass correlation to evaluate the inter-rater consistency (where we treat each run as a rater). The ICC score for police presence is 0.934 (95% CI : [0.921,0.946]), indicating excellent reliability. While the ICC score for protest size is 0.518 (95% CI :[0.463,0.577]), indicating moderate reliability, this suggests that the results are not very reliable. It is quite evident that results are more reliable or consistent when performing simpler tasks than complex ones (i.e., binary versus categorical variable).

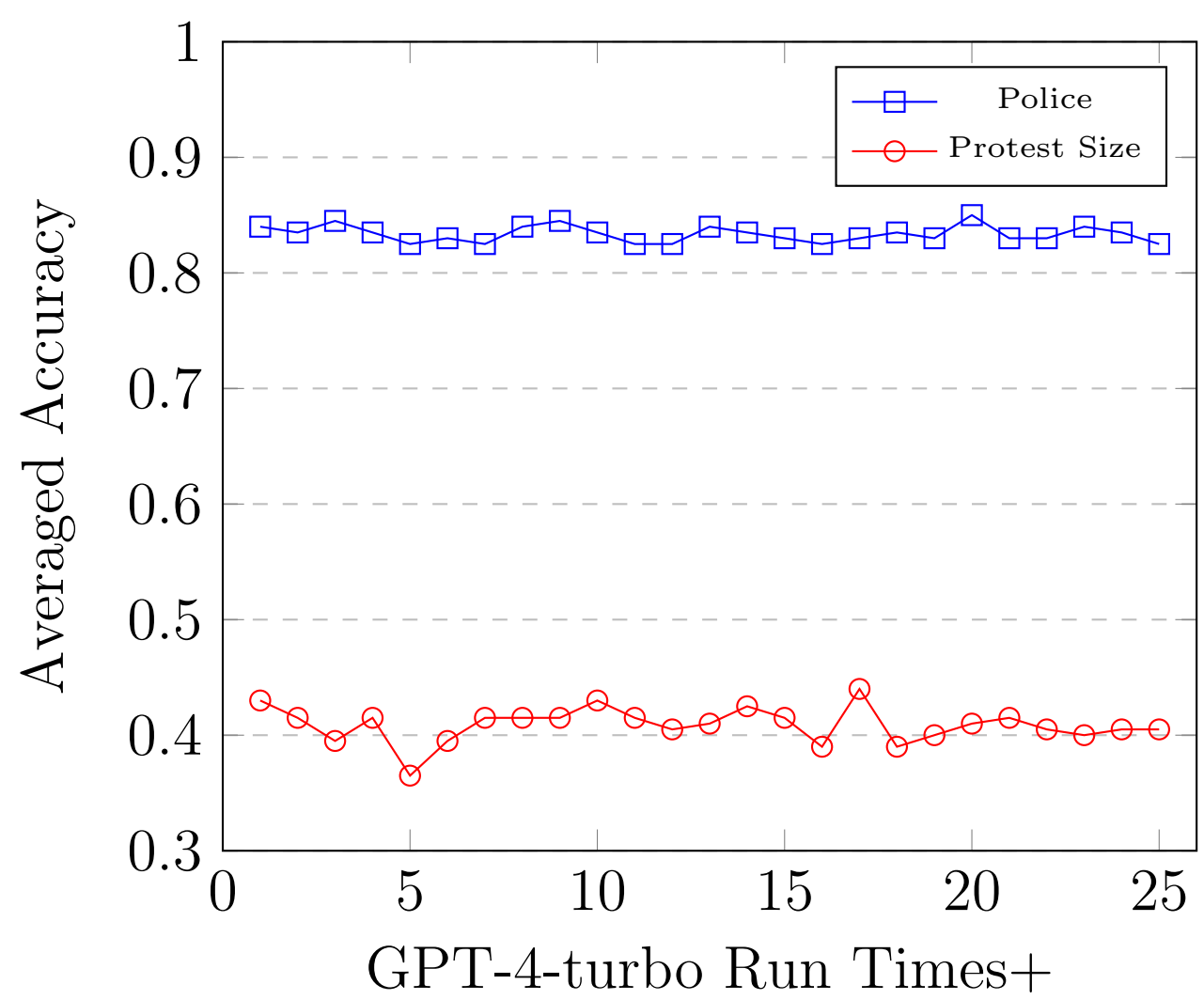

Figure 2: The Consistency of 25 Repeated Outcomes.

## Are LLMs text annotation results replicable?

The potential lack of replicability is another hurdle that lowers social scientists' enthusiasm for adopting LLMs in their research (Barrie, Palaiologou, & Törnberg, 2024). Recent work by Barrie et al. (2024) using 6 datasets covering 12 classification tasks, shows that variations in prompt phrasing can lead to significant differences in model output. This raises an important question: can other researchers replicate LLMs' results using the same methods and data? The answer is complex. Replicability depends on various factors, including prompts, model parameters, model type, and more.

*The alchemy of prompt engineering.* The performance of annotation tasks is highly sensitive to prompt engineering. In this paper, we tested several common prompting strategies, including zero-shot with role play, chain-of-thought (CoT), and tree-of-thought (ToT). Our experiments show that LLM performance varies significantly depending on the prompting strategy used. We find that the simple zero-shot with role play performs well across different tasks. However, for binary classification, zero-shot CoT outperforms other strategies, while for multiclass classification, zero-shot ToT yields better results. This is not to suggest that scholars must adopt CoT or ToT in their research but rather to highlight that different prompting strategies can lead to varying outcomes.

*Model drift.* The rapid development of AI brings the unprecedented challenges to social scientists in tandem with opportunities. Advanced models offer scholars more power to handle multimodal data with improved performance. However, these advancements also introduce uncertainty and replicability issues. For instance, OpenAI offers a series of models, such as gpt-4, gpt-4o, and gpt-3.5, designed for different tasks with varying pricing and speeds. When some of those legacy models become inaccessible to the public, scholars cannot replicate their results. Moreover, model drift occurs when a model's performance degrades over time due to changes in the data distribution or the environment in which it operates, making the model less effective (Vela et al., 2022; Chen, Zaharia, & Zou, 2023). Any change in an upstream AI model's settings introduces less replicability of downstream tasks such as text annotation.

*Personalization and memory.* The alignment of LLMs with human values and norms through reinforcement learning from human feedback has been employed to enhance the safety of LLMs. However, the alignment norms are often dictated by a small group of developers or researchers, leading to the exclusion and under-representation issues. Meanwhile, to better reflect individual preferences, proprietary models such as OpenAI's GPT-4 can be tailored to individuals through memory or personalization feature. Personalized LLMs can reduce cultural hegemony and improve usefulness, but also introduce risks pertaining to privacy infringement and bias reinforcement (Kirk, Vidgen, Röttger, & Hale, 2024). Although personalization is the new frontier in LLM advancement, it poses significant challenges for social scientists, particularly concerning replicability. This is especially problematic when researchers use proprietary models with memory enabled to perform coding tasks.

## Are LLMs text annotation results transparent?

The use of LLMs - particularly proprietary models - often operates as a black box. Scholars have warned of the dangers associated with using proprietary LLMs in social science research, in large part due to the transparency issues (Bender, Gebru, McMillan-Major, & Shmitchell, 2021; Ollion, Shen, Macanovic, & Chatelain, 2024). One strategy for social scientists to improve the coding transparency is to ask LLMs to output their reasoning steps which might help us understand the underlying logic. In so doing, human evaluators can leverage the reasoning outputs for cross-validation. Below is an example using zero-shot CoT to ask gpt-4-turbo to determine whether an article discusses peace movement. gpt-4-turbo reasons output the following steps:

1. *The article mentions the arrest of eleven people, including antiwar activist Philip Berrigan, who were part of a demonstration against war. They scaled a fence at the White*

*House and threw red ink into a fountain as a form of protest.*

2. *The demonstrators are described as being affiliated with an antiwar group, specifically named the D.C. Coalition Against the U.S. in the Middle East.*
3. *This indicates that the protest was related to a peace movement, as it involved a group actively opposing war in the Middle East.*

These steps detail how gpt-4-turbo evaluates whether an article discusses a peace movement based on several key pieces of information. However, scholars should approach the step-by-step reasoning results with caution due to the risk of model hallucination. LLMs may simply generate results conforming to the prompt instead of genuinely reasoning based on the content they have processed (B. Wang et al., 2023).

To investigate this further, we selected 100 news articles and prompted GPT-4 and Llama 3.3 to use zero-shot CoT to generate step-by-step reasoning for two variables - police presence and protest size. We then manually examined whether these outputs were factually correct. Based on our human evaluation, GPT-4 correctly reasoned through all 100 cases for police presence without any signs of hallucination when assessed against the materials extracted from the articles. When classifying protest size, GPT-4 made errors in its reasoning. From the reasoning we can see GPT-4 successfully identified the key information necessary to address the question and assigned it to the correct category based on the prompt instructions in most cases, but occasional misinterpretations or inconsistencies were observed. Analyzing these reasoning errors can be informative, as it helps us understand the sources of incorrect outputs and refine the prompts to address recurring error patterns. Below are some examples of GPT-4's wrong reasoning:

Examples of gpt-4-turbo' Wrong Reasoning

**Example 1**

**Output:** *3 <step by step reasoning>.*

*1. The article mentions '32 Cubans were entering their second week without food' as part of a hunger strike, indicating the number of participants specifically in that action.*

*2. The categorization scale provided dictates: 1 people is a small handful, 10 is considered a group or committee, 50 is a large gathering, 100 is hundreds/mass/mob,*
*1,000 is thousands, and 10,000 or more is tens of thousands.*

*3. Given that 32 people are participating in the hunger strike, this number falls within the 'Group, committee (10 people)' category on the provided scale.*

*4. Therefore, the numeric category for the number of participants in this event, based on the information given in the article, is 2.*

**Where the error lies:** *The categories are inconsistent when reporting it (3) and in the reasoning (2).*

**Example 2**

**Output:** *2 <step by step reasoning>.*

*1. The article specifically mentions 'five Lane College students' as the participants of the sit-in.*

*2. There is no mention of a larger group directly participating with the five students in the sit-in, suggesting the event didn't escalate beyond those directly involved.*

*3. The description does not indicate a larger crowd of protesters but does mention a crowd of white persons reacting to the sit-in, however, the focus is on the protesters.*

*4. Given this information, the number of participants involved in the actual protest (the sit-in) falls into the 'Group, committee (10 people)' category, specifically because the documented participants number exactly five, which is within the range for category 2.*

**Where the error lies:** *First, the key information is wrong: 'five Lane College students' were just part of the sit-in crowd. Second, let's assume the key information has been correctly identified, the category should be 1 (1 - 9 people) instead of 2 (10 - 49 people).*

**Example 3**

**Output:** *1 <step by step reasoning>.*

*The article provided details about a lawsuit filed by the N.A.A.C.P. and the New York Open Housing Center against The New York Times regarding racially discriminatory housing advertisements.*

*However, there was no specific information about a protest event or the number of participants involved in any related gathering.*

*Therefore, based on the information given, the article does not describe a protest with participants that can be categorized using the provided scale.*

**Where the error lies:** *Correctly identified the key information, but the category is wrong.*

We found errors in Llama 3.3 outputs for both variables. Compared to GPT-4, although Llama 3.3 makes other mistakes in reasoning, it has a stronger tendency to repeatedly make the same type of error in both tasks - the inference goes too far. While LLMs are powerful because they can effectively leverage their pretrained knowledge to tackle new tasks, Llama 3.3 tends to override the input (in this case the prompt and the news articles) with its prior knowledge. Below are some examples of Llama 3.3' wrong reasoning:

**Examples of Llama' Wrong Reasoning - Part 1**

**Example 1 - Presence of Police**

**Output:** *1-Yes; Here is my step-by-step reasoning:*
** The article describes an event where David Duke takes the oath of office as a member of the Louisiana Legislature.*
** The event took place in the House chamber, which suggests that it was held at the Louisiana State Capitol building or possibly another government building.*
** Later in the article, it mentions "lawmakers" and "Representative C.D. Jones", who were present at the event and reacted to Mr. Duke's swearing-in.*
** From this context, I infer that police officers would likely be present at such an event, especially considering the controversy surrounding David Duke's appointment.*
** Therefore, I conclude that police were indeed present at the event mentioned in the news article, which is why my answer is 1-Yes.*
**Where the error lies:** *The news article does not mention the presence of police, but Llama infers from the context that police officers would likely be present at such event.*

**Example 2 - Presence of Police**

**Output:** *1-Yes; Step by step reasoning:*
*The article mentions that the student editors have protested the rule by handing out their publications across campus, which could prompt the university to disband their groups. This implies that there may be some interaction or presence of police at this event.*
**Where the error lies:** *The news article does not mention the presence of police, but Llama infers from the context that police officers would likely be involved in such situation.*

**Example 3 - Protest Size**

**Output:** *6. The article does not explicitly mention the number of participants at a specific event. However, it discusses court cases and legal proceedings involving judges, attorneys, and plaintiffs. Since there is no direct reference to an event with a specific number of attendees, we must look for any indirect clues or related information.*
*...(Part of the reasoning has been omitted for brevity.)*
*6. The article discusses legal proceedings and their implications rather than a specific gathering or event, leading to confusion in applying the participant categorization scale directly; however, considering the broad impact of the law on the population, it might imply involvement or effects on a larger scale, though not as traditionally defined by event attendance.*
**Where the error lies:** *The news article does not specify the size of the protest, but Llama first assumes the broad impact of the law and then infers the size based on this assumption.*

**Examples of Llama' Wrong Reasoning - Part 2**

**Example 4 - Protest Size**
**Output:** *6. The article mentions specific individuals involved in the event, such as Frank Townsend and Joseph A. Colombo Sr., but does not provide a direct count of participants at the scene of the bombing.*
*However, considering that the explosion occurred outside an apartment building and the involvement of a group claiming responsibility (the Black Revolutionary Assault Team), it's reasonable to infer that there could have been numerous people in the vicinity or associated with the groups involved, potentially exceeding tens of thousands when considering all individuals indirectly affected or associated (residents of the area, members of the black group, police, etc.).*
*Since we cannot accurately quantify participants directly from the information given and must consider the broader context, the event's impact suggests a large-scale involvement or interest.*
**Where the error lies:** *The news article does not specify the size of the protest, but Llama first assumes the broad impact of the event and then infers the size based on this assumption.*

**Example 5 - Protest Size**
**Output:** *4 step by step reasoning:*
*The article mentions that there are 42 Cuban refugees participating in the hunger strike, and they are among 244 refugees from the boat lift who have been ordered detained.*
*Since 42 is between 10 and 999, but more specifically between 10 and 99 would categorize it as a "Large gathering" however since 42 is closer to the lower end of that range, and there's another larger group mentioned (244), we should consider which number is being asked about - participants in the event, which refers to those on hunger strike.*
*Therefore, given that 42 falls into the range of 10 - 999 people but doesn't reach the hundreds, we can categorize it more precisely as "Hundreds, mass, mob" because it's closer to this category due to its relatively small size compared to other possible large gatherings, however the number is still below a hundred.*
**Where the error lies:** *Correctly identified the key information, but the category should be 2 (10 - 49 people) instead of 4 (100 - 999 people).*

Another widely shared concern among social scientists using LLMs for text annotation is the confidence of model output. How confident is the LLM in its classification? When proprietary LLMs do not provide the distribution of token probability, it poses greater epistemic risks, as scholars are not informed about the confidence of the results. OpenAI GPT-4 provides a feature that allows scholars to access the *logprob* parameter to evaluate the confidence.

Logprob refers to the logarithmic probability of output tokens, indicating the likelihood of each token occurring in a sequence given its preceding context. Simply put, a logprob is $\log(p)$, where $p$ is the probability of a token appearing at a specific position in the sequence based on the preceding tokens. Logprobs fall within the range $s \in (-\infty, 0]$, where 0 corresponds to 100% certainty, or absolute confidence that the token will appear next. Higher log probabilities suggest a greater likelihood of the token in that context(Tian et al., 2023). While a higher logprob indicates higher model confidence, it does not guarantee greater prediction accuracy; rather, it measures the model's internal certainty based on contextual understanding. For researchers, log probabilities offer valuable insights into the model's confidence level for each predicted token, providing an additional diagnostic tool for interpreting model outputs.

With logprob, we can calculate another metric of uncertainty: perplexity. Perplexity is determined by exponentiating the negative average of the logprobs (Gonen, Iyer, Blevins, Smith, & Zettlemoyer, 2022). Generally, a higher perplexity indicates greater uncertainty, while a lower perplexity suggests higher confidence. Thus, perplexity can be used to evaluate the result of an individual model run and to compare the relative confidence of results across different model runs. Below is the formula for perplexity, where the probability of a sequence $W$, composed of the words $w_1, w_2, \ldots, w_n$, is expressed as follows:

$$\text{Perplexity}(P) = 2^{-\frac{1}{N}\sum_{i=1}^{N}\log_2 P(w_i|w_1,w_2,\ldots,w_{i-1})}$$

As an example, we use *police presence* to demonstrate how token probabilities can be derived using the *logprobs* parameter. We used the following prompt, and set *logprobs* to True when utilizing OpenAI's API:

*You need to determine if it mentions police at the event in the following article: <full text>. Please structure your response (1-Yes; 0-No) as follows: <numeric answer>.*

Figure 3 shows the extracted predicted token probability (i.e., classification results). The majority of cases reported probabilities exceeding 95%. For clarity, we only report the 137 cases with probabilities below 95%. These cases with a low level of probability or confidence often warrant further investigation to ensure accuracy.

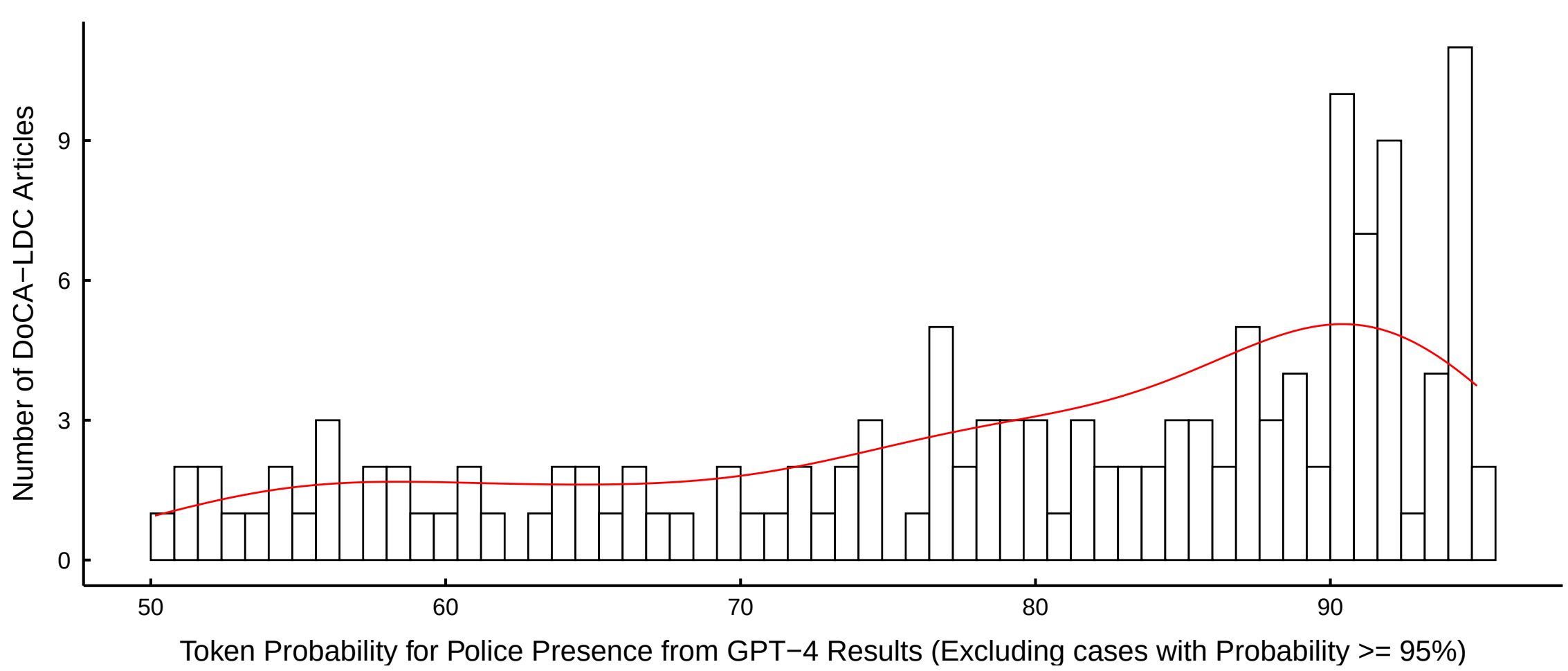

Figure 3: Histogram of Reported Probability for Police Presence

# The Future of Text Annotation with LLMs

### Ascertaining the role of LLMs for text annotation in social science research

LLMs can serve as invaluable tools in social science research, functioning either as primary coders or secondary coding assistants. These models are particularly advantageous for handling large volumes of text that would otherwise require significant human effort. First, scholars can leverage LLMs to perform intermediate steps, such as generating text summarizations, which can then be accessed by human coders for further analysis in downstream tasks. This application is particularly useful when dealing with substantial amounts of text. To obtain detailed coding results with reasoning, researchers need to craft specific prompts, such as Chain of Thought (CoT), to effectively guide the LLMs. Second, in scenarios where high accuracy is paramount, the dataset is relatively small, and the budget allows for hiring additional research assistants, LLMs typically assume a secondary assistant role to support human coders in their decision-making processes. In scenarios where the error tolerance is relatively higher, the dataset is large, and the coding task is not complicated, LLMs can function as primary coders. In other words, scholars fully depend on these models to complete coding tasks such as classification and regression problems. Last, needless to say, scholars can use LLMs as primary coders to generate labeled training datasets in order to fine-tune their own LLMs on large-scale datasets. Given the divergence in the workflow, scholars need to ascertain the role of LLMs in their specific coding tasks.

### Enhancing the quality of text annotation with prompt engineering

Our experiments indicate that the quality of LLMs annotation outputs is highly sensitive to the specific prompts used. Consequently, scholars need to pay particular attention to prompt engineering—finding the best-performing prompt for their specific coding tasks. For LLMs serving as secondary coding assistants, we recommend that researchers experiment with the CoT prompting strategy. This approach guides LLMs to produce step-by-step reasoning, providing scholars with detailed information for human evaluation and decision-making in coding tasks. For LLMs that serve as primary coders, it is crucial for scholars to create an development dataset to systematically test the performance (i.e., both accuracy and reliability) of different prompting strategies. This testing phase ensures that the chosen prompt candidate is valid and reliable before its application to the entire dataset. We also propose that scholars can report on the log probability of predicted tokens and select the cases with low confidence to diagnose the potential issues leading to low accuracy. In this paper, we only showed several common prompting strategies including zero-shot, role play, chain of thought, and tree of thought. Other prompting strategies, such as in-context learning with few examples and graph of thoughts, are not covered in this article, but they are also important and effective. Notably, with sufficient budget, scholars can also fine-tune LLMs for specific downstream tasks with their own training datasets.

### Decomposing complex tasks into simple ones such as binary classification

We tested several common coding tasks in social science research, particularly in social movement studies. Our experiments show that GPT-4 performs better in simple binary classification task than multilabel and multiclass classification tasks. As the task complexity increases, LLMs' performance becomes worse. We recommend that scholars should decompose complex coding tasks into simple ones. For instance, scholars can modify multilabel or multiclass tasks into a series of binary classification.

### Improving the interpretability by outputting the underlying reasoning and coding decision

One of the major concerns of using LLMs in social science coding is their black box nature, which limits interpretability. We propose that scholars can mitigate this issue by tailoring prompts to guide LLMs to generate underlying reasoning for human evaluation. Rather than outputting only the final result, LLMs can be instructed to produce step-by-step reasoning that explains their coding decisions. However, scholars should exercise caution, as the reasoning produced by LLMs may not accurately reflect their internal understanding; instead, it may result from hallucinations or simply conform to the structure and guidance of the prompt (B. Wang et al., 2023). Our human evaluation of 100 cases across two variables reveals differences in the results of the underlying reasoning across tasks and models. For GPT-4, we did not identify any errors when coding police presence in protests; however, errors emerged when coding protest size, as the task complexity increased. In contrast, for Llama 3.3, we observed errors in both tasks, with a tendency to hallucinate - inferring information beyond what was explicitly stated in the text, likely by drawing on its pretrained knowledge. Understanding those errors through reasoning can be helpful, as it helps us identify where the coding decision can go wrong.

### Increasing the replicability by making codes and outputs transparent

To improve the replicability of using LLMs in social science research, we encourage scholars to deposit and share codes in public repository such as GitHub, OSF, and SocArxiv, documenting the specific prompt, the specific model, and the specific parameter setting. If possible, we also call for scholars to share their outputs from LLMs. We acknowledge that scholars should anonymize/remove some of these input/output data that contain sensitive information (Li, Dohan, & Abramson, 2021).

### Establishing repeated researcher audits of the performance of LLMs

Prior research has documented significant performance drift in LLMs over time (Chen, Zaharia, & Zou, 2024), which challenges the assumption that model updates invariably enhance accuracy and underscores the necessity of systematic, ongoing evaluations to monitor the models' performance and behavior. However, existing audit frameworks rarely address the methodological and epistemological concerns of social science research (Jang, Georgiadis, Zhao, & Silavong, 2024). To bridge this gap, we call for the establishment of a collaborative framework in which researchers and journals routinely assess, document, and publish LLMs' performance for social scientific inquiries. Moreover, we advocate for developing more domain-specific benchmarks such as the DoCA project, alongside structured prompts tailored to evaluate models within relevant contexts. Such practices would align LLMs evaluations with the theoretical and empirical priorities of social science research.

**Prioritizing open-source models over proprietary LLMs for data security and responsible use**

Social scientists should practice responsible and ethical use of LLMs in their own research. One of the primary ethical concerns is related to leaking proprietary and confidential data when using LLMs. Social scientists often engage with proprietary and confidential data, such as the New York Times news articles, voter records, interview transcripts, and survey. When researchers use proprietary LLMs from tech companies such as OpenAI and Google, essentially, they have to input their data to the models, which may cause potential data leakage and privacy violation. One common task of social movement scholars is to extract key protest information from news articles. Would inputting these news articles to proprietary LLMs for classification tasks cause data security issue? We propose four suggestions. First, we can opt out of allowing proprietary LLMs to use your uploaded data and usage history for model training or sharing with third parties. Second, qualitative scholars, for example, who interview protesters on sensitive topics, should consider the ethical implications of sharing such data with LLMs for transcription and analysis. In such cases, researchers should follow standard protocols, such as obtaining explicit permission from participants to use LLM tools for analyzing their data. Third, similar to handling survey data, confidential data should be de-identified or anonymized before processing. Finally, an alternative to proprietary LLMs is to prioritize the use of open-source LLMs, such as the Llama models, in social science research.

The advantage of these open-source models is that you can deploy these models using frameworks like LangChain, ollama, or HuggingFace Transformers, and run them on your own computers or remote servers within a secure environment. This way, you do not share your data with external servers when using those models. In this study, we evaluated the performance of the Meta Llama3 model. Specifically, we experimented with two Llama3 models: a 4-bit quantized version of Llama 3.1, known as Llama-3-8B, and Llama-3.3-70B, across three tasks. We applied the same zero-shot prompt technique used for the gpt-4-turbo model at a temperature of 0.2. The Llama-3-8B model is lightweight and fast, enabling us to deploy it on a local computer using the Ollama interface. In contrast, due to its larger size, the Llama-3.3-70B model was deployed on a high-performance computing system.

The completion times for each task using Meta-Llama-3-8B were as follows: 1.6 hours for the binary task assessing police presence, 2.2 hours for the multiclass task determining the number of participants, and 5.5 hours for the multilabel task identifying protest activities. For Llama-3.3-70B, the completion times were 7.2 hours for the binary task, 8 hours for the multiclass task, and 8.7 hours for the multilabel task. The results for both Llama 3.1 and Llama 3.3 are presented in Table 2. It shows that Llama-3.3-70B has better performance than Llama-3-8B. However, when compared to gpt-4-turbo, Llama-3.3-70B has slightly lower performance across all three tasks. Notably, the performance gap between GPT-4 and Llama 3.3 was much smaller for less complex tasks, such as binary and multiclass variables.

| Task Complexity | Llama-3-8B | Llama-3.3-70B |
|---|---|---|
| Binary (Police) | 0.76, 0.57 | 0.77, 0.57 |
| Multiclass (Size) | 0.37, 0.28 | 0.44, 0.43 |
| Multilabel (Activity) | 0.10, - | 0.27, - |

Table 2: Performance of Llama3 Models based on Zero-shot CoT.

# Conclusion and Discussion

LLMs have been applauded for its potential application in social science research including automated text analysis (Ziems et al., 2024; Bail, 2024). But can we replace human coders with LLMs in textual analysis? Our findings indicate both promise and pitfalls. In this paper, we propose a framework to integrate LLMs in social science coding by assuming two roles: LLMs as secondary coding assistants and LLMs as primary coding decision-makers. When LLMs are the secondary coding assistants, human coders can use the information extracted/summarized by prompted LLMs to assist their decision making processes. The epistemic risks associated with this role are relatively low since LLMs perform pretty well in information retrieval and human coders make the final decision. However, when LLMs assume the primary coder role, scholars should be cautious of replacing human coders with LLMs simply because they can make mistakes and reproduce existing biases. For example, they tend to misportray marginalized groups as more like outgroup imitations instead of in-group representations, flatten demographic groups with ignoring within-group heterogeneity, and essentialize identities with amplifying stereotypes (Dillion, Tandon, Gu, & Gray, 2023). To better communicate the epistemic risks associated with using LLMs in social science coding, we propose several practice guidelines, including the before-prompt engineering and after-evaluation pipeline to ensure validity and reliability of the chosen prompt and model, the explainable annotation by outputting underlying reasoning, and improving transparency by sharing codes, prompts, model parameters, and returned outcomes.

When using AI for research, especially proprietary models, one of the biggest concerns is data security and privacy. As social scientists, we must pay close attention to ethical considerations when employing these models in our work. We recommend that scholars prioritize open-source LLMs to ensure data security. Deploying an open-source model locally or on a remote server in a secure environment ensures that no data is shared with external servers. Additionally, researchers should obtain explicit permission from participants when using proprietary LLMs to analyze sensitive data. This paper primarily focuses on the use of LLMs for text annotation in social science research. However, LLMs represent just one type of text-based generative AI with capabilities such as text generation, translation, summarization, and information extraction. Generative AI extends beyond text to produce human-like content in formats such as images, audio, and video (Bail, 2024). We note that, as generative AI advances, it continues to contribute to social science discovery in areas such as simulation-based research (Bail, 2024). At the same time, social science expertise is essential for evaluating and understanding the societal impacts of generative AI (Burrell & Fourcade, 2021; Lei & Kim, 2024). Such insights can, in turn, guide the development of generative AI in ways that mitigate its harmful impact on society (Sloane & Moss, 2019).

# Appendix

## Appendix A

Below are a list of example prompts we used for GPT-4 to code protest size.

**Zero-shot**: You need to extract the number of participants and tell me which category: 1 – Small, few handful (1-9 people); 2 – Group, committee (10-49 people); 3 – Large, gathering (50-99 people); 4 – Hundreds, mass, mob (100-999 people); 5 – Thousands (1000-9999 people); 6 – Tens of thousands (10000 or more people) in the following article: < *fulltext* >. Please structure your numeric response (1-6) as follows: <numeric answer >

**Zero-shot CoT:** You need to extract the number of participants and tell me which category: 1 – Small, few handful (1-9 people); 2 – Group, committee (10-49 people); 3 – Large, gathering (50-99 people); 4 – Hundreds, mass, mob (100-999 people); 5 – Thousands (1000-9999 people); 6 – Tens of thousands (10000 or more people) in the following article: < *fulltext* >. Please structure your numeric response (1-6) as follows: <numeric answer >. Let us think step by step.

**Zero-shot ToT:** Imagine three different social movement experts are answering this question. All experts will write down one step of their thinking, then share it with the group. Then all experts will proceed to the next step, and so on. If any expert realizes they're wrong at any point, they will leave. The question is that You need to extract the number of participants and tell me which category: 1 – Small, few handful (1-9 people); 2 – Group, committee (10-49 people); 3 – Large, gathering (50-99 people); 4 – Hundreds, mass, mob (100-999 people); 5 – Thousands (1000-9999 people); 6 – Tens of thousands (10000 or more people) in the following article: $< fulltext >$. Please structure your numeric response (1-6) as follows: <numeric answer >

## Appendix B

**Dataset:** The benchmark dataset is from the Dynamics of Collective Action (DoCA) project (Earl et al., 2004; Hanna, 2017). This project focuses on collective action in the United States and annotates news articles from the New York Times covering the period 1960-1995. It is a detailed event database containing information at both the article and event levels, such as the size of protests, police involvement, actions, claims, social movement organizations, locations, timing, and more. Further details can be found here: `https://web.stanford.edu/group/collectiveaction/`. As the DoCA dataset does not include the original news articles, we matched it with the annotated NYT corpus (1987-2007) from the Linguistic Data Consortium (LDC) for our analysis.